\documentclass[11pt,a4paper]{article}
\usepackage[hyperref]{acl2018}
\usepackage{times}
\usepackage{latexsym}
\usepackage{graphicx}

\usepackage{url}
\usepackage{booktabs} 
\usepackage{amsmath}
\usepackage{amsfonts}

\aclfinalcopy 
\def\aclpaperid{1182} 

\title{A Multi-sentiment-resource Enhanced Attention Network for Sentiment Classification}

\author{Zeyang Lei$^{1,2}$, Yujiu Yang$^1$, Min Yang$^3$, and Yi Liu$^2$\\
  Graduate School at Shenzhen, Tsinghua University$^1$ \\
  Peking University Shenzhen Institute$^2$ \\
  Shenzhen Institutes of Advanced Technology, Chinese Academy of Sciences$^3$ \\
  {\tt leizy16@mails.tsinghua.edu.cn, yang.yujiu@sz.tsinghua.edu.cn},\\
  {\tt min.yang1129@gmail.com, eeyliu@gmail.com}
  }

\date{}

\begin{document}
\maketitle
\begin{abstract}
Deep learning approaches for sentiment classification do not fully exploit sentiment linguistic knowledge.
In this paper, we propose a Multi-sentiment-resource Enhanced Attention Network (MEAN) to alleviate the problem by integrating three kinds of sentiment linguistic knowledge (e.g., sentiment lexicon, negation words, intensity words) into the deep neural network via attention mechanisms. By using various types of sentiment resources, MEAN utilizes sentiment-relevant information from different representation sub-spaces, which makes it more effective to capture the overall semantics of the sentiment, negation and intensity words for sentiment prediction.
The experimental results demonstrate that MEAN has robust superiority over strong competitors.
\end{abstract}

\section{Introduction}
Sentiment classification is an important task of natural language processing (NLP), aiming to classify the sentiment polarity of a given text as positive, negative, or more fine-grained classes.  It has obtained considerable attention due to its broad applications in natural language processing~\cite{Reviewer1_linguistic,{Reviewer2_CNN}}. Most existing studies set up sentiment classifiers using supervised machine learning approaches, such as support vector machine (SVM)~\cite{SVM}, convolutional neural network (CNN)~\cite{CNN,{CNN+attention}}, long short-term memory (LSTM)~\cite{{LSTM},LR_LSTM}, Tree-LSTM~\cite{05tai2015improved15}, and attention-based methods~\cite{{wanxiaojun_attention},{hierarchical},self-attention,{Reviewer2_attention}}.

Despite the remarkable progress made by the previous work, we argue that sentiment analysis still remains a challenge. Sentiment resources including sentiment lexicon, negation words, intensity words play a crucial role in traditional sentiment classification approaches \cite{lexicon, {Lexicon_tang}}. Despite its usefulness, to date, the sentiment linguistic knowledge has been underutilized in most recent deep neural network models (e.g., CNNs and LSTMs).

In this work, we propose a Multi-sentiment-resource Enhanced Attention Network (MEAN) for sentence-level sentiment classification to integrate many kinds of sentiment linguistic knowledge into deep neural networks via multi-path attention mechanism. Specifically, we first design a coupled word embedding module to model the word representation from character-level and word-level semantics. This can help to capture the morphological information such as prefixes and suffixes of words.
Then, we propose a multi-sentiment-resource attention module to learn more comprehensive and meaningful sentiment-specific sentence representation by using the three types of sentiment resource words as attention sources attending to the context words respectively. In this way, we can attend to different sentiment-relevant information from different representation subspaces implied by different types of sentiment sources and capture the overall semantics of the sentiment, negation and intensity words for sentiment prediction.




The main contributions of this paper are summarized as follows. First, we design a coupled word embedding obtained from character-level embedding and word-level embedding to capture both the character-level morphological information and word-level semantics.
Second, we propose a multi-sentiment-resource attention module to learn more comprehensive sentiment-specific sentence representation from multiply subspaces implied by three kinds of sentiment resources including sentiment lexicon, intensity words, negation words. Finally, the experimental results show that MEAN consistently outperforms competitive methods.

\section{Model}
Our proposed MEAN model consists of three key components: coupled word embedding module, multi-sentiment-resource attention module, sentence classifier module.
In the rest of this section, we will elaborate these three parts in details. The overall framework is shown in Figure 1.
 \begin{figure*}[!ht]
    \centering{\includegraphics[width=\textwidth] {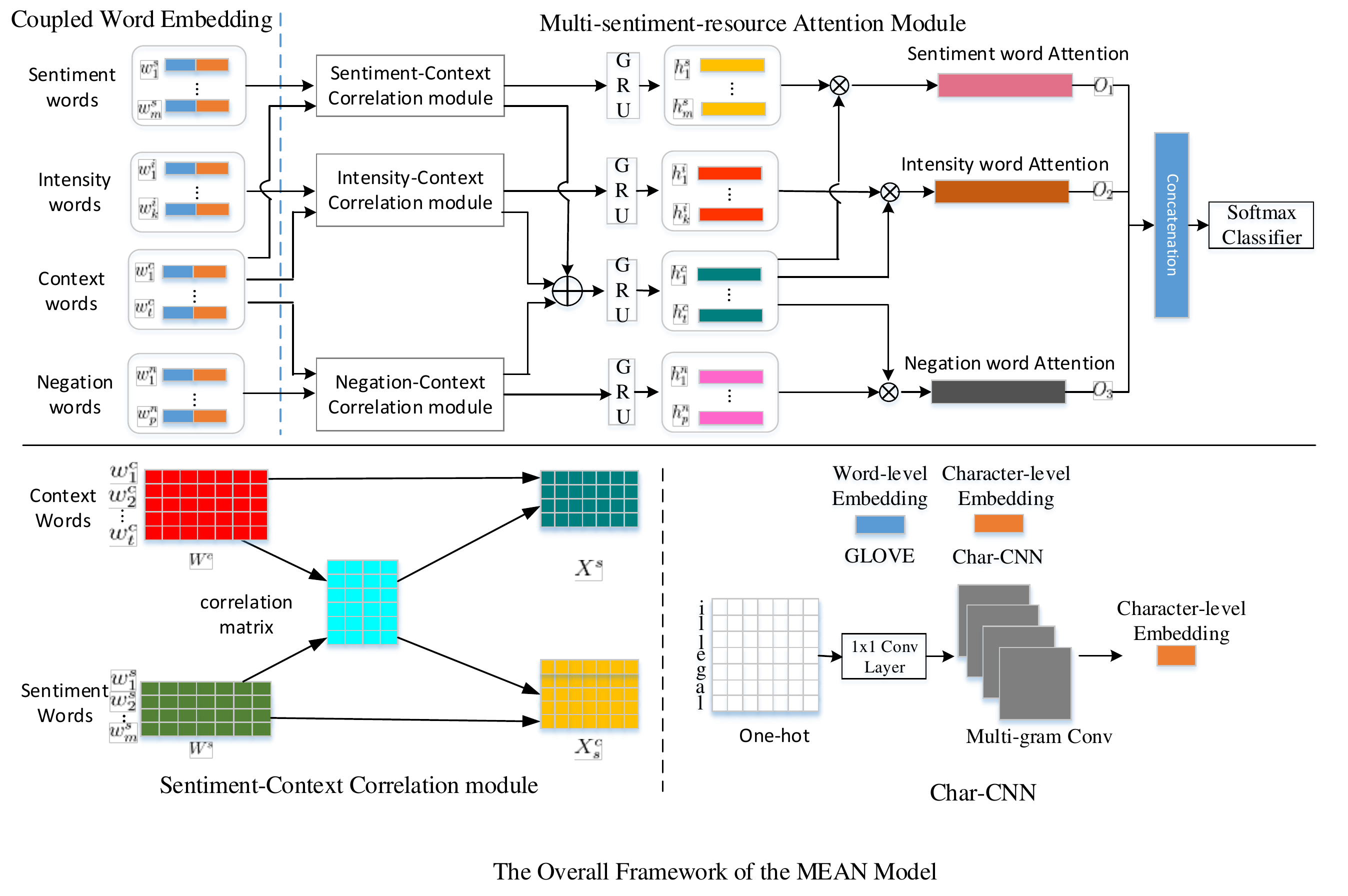}}
    \caption{The Overall Framework of Our Model}
    \label{fig_1}
    \vspace{-15pt}
 \end{figure*}
\subsection{Coupled Word Embedding}
To exploit the sentiment-related morphological information implied by some prefixes and suffixes of words (such as ``Non-", ``In-", ``Im-"), we design a coupled word embedding learned from character-level embedding and word-level embedding. We first design a character-level convolution neural network (Char-CNN) to obtain character-level embedding ~\cite{Charac-CNN}. Different from ~\cite{Charac-CNN}, the designed Char-CNN is a fully convolutional network without max-pooling layer to capture better semantic information in character chunk. Specifically, we first input one-hot-encoding character sequences to a $1\times 1$ convolution layer to enhance the semantic nonlinear representation ability of our model~\cite{FCN}, and the output is then fed into a multi-gram (i.e. different window sizes) convolution layer to capture different local character chunk information.  For word-level embedding, we use pre-trained word vectors, GloVe~\cite{glove}, to map each word to a low-dimensional vector space. Finally, each word is represented as a concatenation of the character-level embedding and word-level embedding. This is performed on the context words and the three types of sentiment resource words \footnote{To be precise, sentiment resource words include sentiment words, negation words and intensity words.}, resulting in four final coupled word embedding matrices: the $W^c=[w_1^c,...,w_t^c] \in \mathbb{R}^{d \times t}$ for context words, the $W^s=[w_1^s,...,w_m^s] \in \mathbb{R}^{d \times m}$ for sentiment words, the $W^i=[w_1^i,...,w_k^i] \in \mathbb{R}^{d \times k}$ for intensity words, the $W^n=[w_1^n,...,w_p^n] \in \mathbb{R}^{d \times p}$ for negation words. Here, $t, m, k, p$ are the length of the corresponding items respectively, and $d$ is the embedding dimension.
Each $W$ is normalized to better calculate the following word correlation.

\subsection{Multi-sentiment-resource Attention Module}
After obtaining the coupled word embedding, we propose a multi-sentiment-resource attention mechanism to help select the crucial sentiment-resource-relevant context words to build the sentiment-specific sentence representation. Concretely, we use the three kinds of sentiment resource words as attention sources to attend to the context words respectively, which is beneficial to capture different sentiment-relevant context words corresponding to different types of sentiment sources. For example, using sentiment words as attention source  attending to the context words helps form the sentiment-word-enhanced sentence representation. Then, we combine the three kinds of sentiment-resource-enhanced sentence representations to learn the final  sentiment-specific sentence representation.
We design three types of attention mechanisms: sentiment attention, intensity attention, negation attention to model the three kinds of sentiment resources, respectively.
In the following, we will elaborate the three types of attention mechanisms in details.

First, inspired by \cite{co-attention}, we expect to establish the word-level relationship between the context words and different kinds of sentiment resource words. To be specific, we define the dot products among the context words and the three kinds of sentiment resource words as correlation matrices. Mathematically, the detailed formulation is described as follows.
\begin{align}
&M^s= {(W^c)}^T \cdot{W^s} \in \mathbb{R}^{t \times m} \\
&M^i= {(W^c)}^T \cdot{W^i} \in \mathbb{R}^{t \times k} \\
&M^n= {(W^c)}^T \cdot{W^n} \in \mathbb{R}^{t \times p}
\end{align}
where $M^s, M^i, M^n$ are the correlation matrices to measure the relationship among the context words and the three kinds of sentiment resource words, representing the relevance between the  context words and the sentiment resource word.

After obtaining the correlation matrices, we can compute the context-word-relevant sentiment word representation matrix $X^s$, the context-word-relevant intensity word representation matrix $X^i$, the context-word-relevant negation word representation matrix $X^n$ by the dot products among the context words and different types of corresponding correlation matrices. Meanwhile,  we can also obtain the sentiment-word-relevant context word representation matrix $X^c_s$  by the dot product between the correlation matrix $M^s$ and the sentiment words $W^s$, the intensity-word-relevant context word representation matrix $X^c_i$ by the dot product between the intensity words $W^i$ and the correlation matrix $M^i$, the negation-word-relevant context word representation matrix $X^c_n$ by the dot product between the negation words $W^n$ and the correlation matrix $M^n$. The detailed formulas are presented as follows:
\begin{align}
&X^s=W^cM^s, X^c_s=W^s(M^s)^T \\
&X^i=W^cM^i, X^c_i=W^i(M^i)^T \\
&X^n=W^cM^n, X^c_n=W^n(M^n)^T
\end{align}

The final enhanced context word representation matrix is computed as:
\begin{equation}
X^c=X^c_s+X^c_i+X^c_n.
\end{equation}

Next, we employ four independent GRU networks~\cite{GRU} to encode hidden states of the context words and the three types of sentiment resource words, respectively.
Formally, given the word embedding $X^c, X^s, X^i, X^n$, the hidden state matrices $H^c, H^s, H^i, H^n$ can be obtained as follows:
\begin{align}
&H^c=GRU(X^c) \\
&H^s=GRU(X^s) \\
&H^i=GRU(X^i) \\
&H^n=GRU(X^n)
\end{align}

After obtaining the hidden state matrices, the sentiment-word-enhanced sentence representation  $o_1$  can be computed as:
\begin{align}
&o_1=\sum_{i=1}^t \alpha_i h^c_i, q^s=\sum_{i=1}^m h_i^s/m \\
&\beta([h_i^c;q_s])=u_{s}^T tanh(W_{s}[h_i^c;q_s]) \\
&\alpha_i = \frac{exp(\beta([h_i^c;q_s]))}{\sum_{i=1}^{t}exp(\beta([h_i^c;q_s]))}
\end{align}
where  $q^s$ denotes the mean-pooling operation towards $H^s$, $\beta$ is the attention function that calculates the importance of the $i$-th word $h_i^c$ in the context and $\alpha_i$ indicates the importance of the $i$-th word in the context, $u_{s}$ and $W_{s}$ are learnable parameters.

Similarly, with the hidden states $H^i$ and $H^n$ for  the intensity words and the negation words as attention sources, we can obtain the intensity-word-enhanced sentence representation $o_2$ and the negation-word-enhanced sentence representation $o_3$.
The final comprehensive sentiment-specific sentence representation $\mathbf{\tilde{o}}$ is the composition of the above three sentiment-resource-specific sentence representations $o_1, o_2, o_3$:
\begin{equation}
\mathbf{\tilde{o}}=[o_1, o_2, o_3]
\end{equation}
\subsection{Sentence Classifier}
After obtaining the final sentence representation $\mathbf{\tilde{o}}$, we feed it to a softmax layer to predict the sentiment label distribution of a sentence:
\begin{equation}
\hat{y}=\frac{exp(\tilde{W_o}^T{{\mathbf{\tilde{o}}}}+\tilde{b_o})}{\sum_{i=1}^Cexp(\tilde{W_o}^T{{\mathbf{\tilde{o}}}}+\tilde{b_o})}
\end{equation}
where $\hat{y}$ is the predicted sentiment distribution of the sentence, C is the number of sentiment labels, $\tilde{W_o}$ and $\tilde{b_o}$ are parameters to be learned.

For model training, our goal is to minimize the cross entropy  between the ground truth and predicted results for all sentences. Meanwhile, in order to
avoid overfitting, we use dropout strategy to randomly omit parts of the parameters on each training case. Inspired by \cite{self-attention}, we also design a penalization term to ensure the diversity of semantics from different sentiment-resource-specific sentence representations, which reduces information redundancy from different sentiment resources attention. Specifically, the final loss function is presented as follows:
\begin{align}
L(\hat{y},y)=&-\sum_{i=1}^N\sum_{j=1}^Cy_i^jlog(\hat{y}_i^j)+\lambda (\sum_{\theta \in \Theta}\theta^2) \\
              &+\mu ||\tilde{O}\tilde{O}^T-\psi I||_F^2 \nonumber \\
\tilde{O}=&[o_1; o_2; o_3]
\end{align}
where $y_i^j$ is the target sentiment distribution of the sentence, $\hat{y}_i^j$ is the prediction probabilities, $\theta $ denotes each parameter to be regularized, $\Theta $ is parameter set, $\lambda $ is the coefficient for $L_2 $ regularization, $\mu$ is a hyper-parameter to balance the three terms, $\psi$ is the weight parameter, $I$ denotes the the identity matrix and $||.||_F$ denotes the Frobenius norm of a matrix. Here, the first two terms of the loss function are cross-entropy function of the predicted and true distributions and $L_2 $ regularization respectively, and the final term is a penalization term to encourage the diversity of sentiment sources.
\section{Experiments}
\subsection{Datasets and Sentiment Resources} Movie Review (MR){\color{blue}\footnote{\url{http://www.cs.cornell.edu/people/pabo/movie-review-data/}}} and Stanford Sentiment Treebank (SST){\color{blue}\footnote{\url{https://nlp.stanford.edu/sentiment/}we train the model on both phrases and sentences but only test on sentences}} are used to evaluate our model. MR dataset has 5,331 positive samples  and 5,331 negative samples. We adopt the same data split as in \cite{LR_LSTM}.  SST consists of  8,545 training samples, 1,101 validation samples, 2210 test samples.  Each sample is marked as very negative, negative, neutral, positive, or very positive. Sentiment lexicon combines the sentiment words from both \cite{LR_LSTM} and \cite{Liubing}, resulting in 10,899 sentiment words in total. We collect negation and intensity words  manually as the number of these words is limited.

\subsection{Baselines}
In order to comprehensively evaluate the performance of our model, we list several baselines for sentence-level sentiment classification.

\textbf{RNTN}: Recursive Tensor Neural Network \cite{04socher2013recursive13} is used to model correlations between different dimensions of child nodes¡¯ vectors.

\textbf{LSTM/Bi-LSTM}: \newcite{DBLP:journals/corr/ChoMGBSB14} employs Long Short-Term Memory and the bidirectional variant to capture sequential information.

\textbf{Tree-LSTM}: Memory cells was introduced by Tree-Structured Long Short-Term Memory \cite{05tai2015improved15} and gates into tree-structured neural network, which is beneficial to capture semantic relatedness by parsing syntax trees.

\textbf{CNN}: Convolutional Neural Networks \cite{CNN} is applied to generate task-specific sentence representation.

\textbf{NCSL}: \newcite{08teng2016context16} designs a Neural Context-Sensitive Lexicon (NSCL) to obtain prior sentiment scores of words in the sentence.

\textbf{LR-Bi-LSTM}: \newcite{LR_LSTM} imposes linguistic roles into neural networks by applying linguistic regularization on intermediate outputs with KL divergence.

\textbf{Self-attention}: \newcite{self-attention} proposes a self-attention mechanism to learn structured sentence embedding.

\textbf{ID-LSTM}: \cite{AAAI} uses reinforcement learning to learn structured sentence representation for sentiment classification.

\subsection{Implementation Details} In our experiments, the dimensions of character-level embedding and word embedding (GloVe) are both set to 300. Kernel sizes of multi-gram convolution for Char-CNN are set to 2, 3, respectively.  All the weight matrices are initialized as random orthogonal matrices, and  we set all the bias vectors as zero vectors.  We optimize the proposed model with  RMSprop  algorithm, using mini-batch training. The size of mini-batch is 60.  The dropout rate is 0.5, and the coefficient $\lambda$ of $L_2$ normalization is set to $10^{-5}$. $\mu$ is set to $10^{-4}$. $\psi$ is set to $0.9$. When there are not sentiment resource words in the sentences,  all the context words are treated as sentiment resource words to implement the multi-path self-attention strategy.

\subsection{Experiment Results} In our experiments, to be consistent with the recent baseline methods, we adopt classification accuracy as evaluation metric. We summarize the experimental results in Table {\color{red}\ref{table1}}. Our model has robust superiority over competitors and sets state-of-the-art on MR and SST datasets.  First, our model brings a substantial improvement over the methods that do not leverage sentiment linguistic knowledge (e.g., RNTN, LSTM, BiLSTM, CNN and ID-LSTM) on both datasets. This verifies the effectiveness of leveraging sentiment linguistic resource with the deep learning algorithms.
Second, our model also  consistently outperforms LR-Bi-LSTM which integrates linguistic roles of sentiment, negation and intensity words into neural networks via the linguistic regularization.  For example, our model achieves $2.4\%$ improvements over the MR dataset and $0.8\%$ improvements over the SST dataset compared to LR-Bi-LSTM. This is because that MEAN designs attention mechanisms to leverage sentiment resources efficiently, which utilizes the interactive information between context words and sentiment resource words.

In order to analyze the effectiveness of each component of MEAN, we also report the ablation test in terms of discarding character-level embedding (denoted as MEAN w/o CharCNN) and sentiment words/negation words/intensity words (denoted as MEAN w/o sentiment words/negation words/intensity words). All the tested factors contribute greatly to the improvement of the MEAN. In particular, the accuracy decreases sharply when discarding the sentiment words. This is within our expectation since sentiment words are vital when classifying the polarity of the sentences.

\begin{table}[!ht]
		\footnotesize
			\centering
			\begin{tabular}{ccl}
			\toprule
			  Methods &  MR &  SST \\
            \midrule
              RNTN  & 75.9\%\# & 45.7\%  \\
              LSTM  & 77.4\%\# & 46.4\%  \\
              BiLSTM & 79.3\%\# & 49.1\%  \\
             Tree-LSTM  & 80.7\%\# & 51.0\%  \\
              CNN  & 81.5\% & 48.0\%  \\
             NSCL  & 82.9\% & 51.1\%  \\
             LR-Bi-LSTM  & 82.1\% & 50.6\%  \\
             Self-attention & 81.7\%* & 48.9\%* \\
             ID-LSTM &81.6\% & 50.0\% \\
             \textbf{MEAN(our model)}   & {\bf{84.5}}\% & {\bf{51.4}}\% \\
             \hline
             MEAN w/o CharCNN &83.2\% & 50.0\%   \\
             \hline
             MEAN w/o sentiment words & 82.1\% & 48.4\% \\
             MEAN w/o negation words & 82.9\% & 49.5\% \\
             MEAN w/o intensity words & 83.5\% & 49.3\% \\
             \bottomrule
			\end{tabular}
            \caption{Evaluation results. The best result for each dataset is in bold. The result marked with \# are retrieved from \cite{LR_LSTM}, and the results marked with * denote the results are obtained by our implementation.}
            \label{table1}
\end{table}

\section{Conclusion}
In this paper, we propose a novel Multi-sentiment-resource Enhanced Attention Network (MEAN) to enhance the performance of sentence-level sentiment analysis, which integrates the sentiment linguistic knowledge into the deep neural network.
\section*{Acknowledgements}
We thank all anonymous reviewers for their valuable comments. The corresponding author is Yujiu Yang. This work was supported in part by the Research Fund for the development of strategic emerging industries by ShenZhen city (No.JCYJ20160301151844537 and No. JCYJ20160331104524983).

\bibliographystyle{acl_natbib}
\bibliography{acl2018}

\end{document}